\documentclass{article}

\usepackage{arxiv}

\usepackage{graphicx} % Required for inserting images
\usepackage{amsmath}
\usepackage{algorithm}
\usepackage[noend]{algpseudocode}
\usepackage{multirow}
\usepackage{makecell}
\usepackage{booktabs}
\usepackage{url} 

\usepackage[framemethod=default]{mdframed}
\newmdenv[innerlinewidth=0.5pt, roundcorner=4pt,linecolor=blue,innerleftmargin=6pt,
innerrightmargin=6pt,innertopmargin=6pt,innerbottommargin=6pt]{mybox}

\title{Towards Continual Visual Anomaly Detection\\ in the Medical Domain}

\author{
    Manuel Barusco \\ 
    University of Padova, Padova, Italy \\ 
    \texttt{manuel.barusco@phd.unipd.it} \\ \And
    Francesco Borsatti \\ 
    University of Padova, Padova, Italy \\ 
    \texttt{francesco.borsatti.1@phd.unipd.it} \\ \And
    Nicola Beda \\ 
    University of Padova, Padova, Italy \\ 
    \texttt{nicola.beda@studenti.unipd.it} \\ \And
    Davide Dalle Pezze \\ 
    University of Padova, Padova, Italy \\ 
    \texttt{davide.dallepezze@unipd.it} \\ \And
    Gian Antonio Susto \\ 
    University of Padova, Padova, Italy \\
    \texttt{gianantonio.susto@unipd.it} \\
}

\begin{document}

\maketitle

\begin{abstract}
Visual Anomaly Detection (VAD) seeks to identify abnormal images and precisely localize the corresponding anomalous regions, relying solely on normal data during training.
This approach has proven essential in domains such as manufacturing and, more recently, in the medical field, where accurate and explainable detection is critical. 
Despite its importance, the impact of evolving input data distributions over time has received limited attention, 
even though such changes can significantly degrade model performance.
In particular, given the dynamic and evolving nature of medical imaging data, Continual Learning (CL) provides a natural and effective framework to incrementally adapt models while preserving previously acquired knowledge.
This study explores for the first time the application of VAD models in a CL scenario for the medical field.
In this work, we utilize a CL version of the well-established PatchCore model, called PatchCoreCL, and evaluate its performance using BMAD, a real-world medical imaging dataset with both image-level and pixel-level annotations.
Our results demonstrate that PatchCoreCL is an effective solution, achieving performance comparable to the task-specific models, with a forgetting value less than a 1\%, highlighting the feasibility and potential of CL for adaptive VAD in medical imaging.
\end{abstract}


\begin{mybox}
\textbf{Correct reference for this work}:
\vspace{0.2cm}
\begin{small}
\noindent Manuel Barusco, Francesco Borsatti, Nicola Beda, Davide Dalle Pezze, and Gian Antonio Susto. ``Towards Continual Visual Anomaly Detection in the Medical Domain''. (2025). \textit{In Proceedings of the International Workshop on Personalized Incremental Learning in Medicine}. 
\noindent \url{https://doi.org/10.1145/3746259} 
\end{small}
\end{mybox}

\section{Introduction}

Visual Anomaly Detection (VAD) is a critical and important task in Computer Vision and Machine Learning that aims at identifying anomalous patterns in the input images that deviate from the normal distributions. In VAD, the data collection is very critical; indeed, normal data is abundant, and anomalies are very rare and difficult to collect. Because of this, VAD models are trained in a totally unsupervised way by considering only normal data, while in inference, the models are tested on normal and anomalous data. In order to provide explainability of the predictions, the models must return an image-level score but also a segmentation map that pinpoints the anomalous parts of the image. VAD models are employed in a variety of fields spanning from manufacturing \cite{mvtec} to surveillance \cite{surv_vad} and the medical domain \cite{bao2023bmad}.
\\
In the medical domain, VAD is very delicate, and anomalous patterns may indicate important pathological conditions; this underlines the importance of developing accurate and fast-adapting models in this field.
VAD methods have shown strong performance on static scenarios, where a separate model is trained for each considered domain or category. However, VAD models, like every other machine learning model, are prone to Catastrophic Forgetting: when learning a new task, where a task can represent a new category or a new data domain, the model forgets the previous ones.
\\
Continual learning (CL) is a branch of Machine Learning that enables models to incrementally learn and adapt to new tasks without forgetting the previous ones and without needing a complete retraining of the model on new and old data simultaneously (which may be impossible or too demanding).
\\
In this work, we study for the first time the Continual Visual Anomaly Detection task using the BMAD (Benchmarks for Medical Anomaly Detection) \cite{bao2023bmad} dataset, a diverse collection of medical imaging datasets spanning multiple medical domains as anatomical regions and image acquisition modalities. 
In this work, we adopt PatchCore \cite{patchcore} as Visual Anomaly Detection method, a well-established model for industrial anomaly detection, able to output both image-level anomaly scores and interpretable heatmaps that localize the anomalies in the pixel space. 
In particular, we test PatchCoreCL, a variant of PatchCore proposed for the continual learning setting \cite{bugarin2024unveiling}.
% Furthermore, based on the previous results on the MVTec dataset \cite{bugarin2024unveiling}, this model showed promising results on the Continual Anomaly Detection task.
% Our focus is on evaluating PatchCoreCL, a variant of PatchCore for the continual learning setting.
% performance in this setting by proposing an adaptation to the continual learning scenario.
\\
In summary, the main contributions of this paper are:
\begin{itemize}
    \item We explore for the first time Continual Visual Anomaly Detection setting for the medical field, where new domains are incrementally presented to the VAD model.
    \item Our results are tested on a CL scenario created from the well-established BMAD dataset and combined with PatchCoreCL (a CL variant of the well-known PatchCore model), obtaining a robust baseline for future research.
    \item Our results demonstrate that PatchCoreCL is an effective solution, achieving performance comparable to the task-specific models, demonstrating the feasibility for dynamic medical imaging.
\end{itemize}

Moreover, to promote further research in this new scenario, we provide the code used for the experiments, available on the MoViAD library \cite{moviad}. 
The paper is structured as follows: Section \ref{sec:related} provides an overview of the related works in the fields of VAD and CL, Section \ref{sec:method} describes the details about PatchCore and its continual adaptation, while \ref{sec:exp_setting} outlines the experimental setting. Section \ref{sec:metrics} introduces the considered VAD and CL metrics, and Section \ref{sec:results} comments on the obtained results. In the end, Section \ref{sec:conclusions} draws the conclusions of this work and provides some interesting future work directions. 

\begin{figure}[!th]
  \centering
  \includegraphics[width=0.9\textwidth]{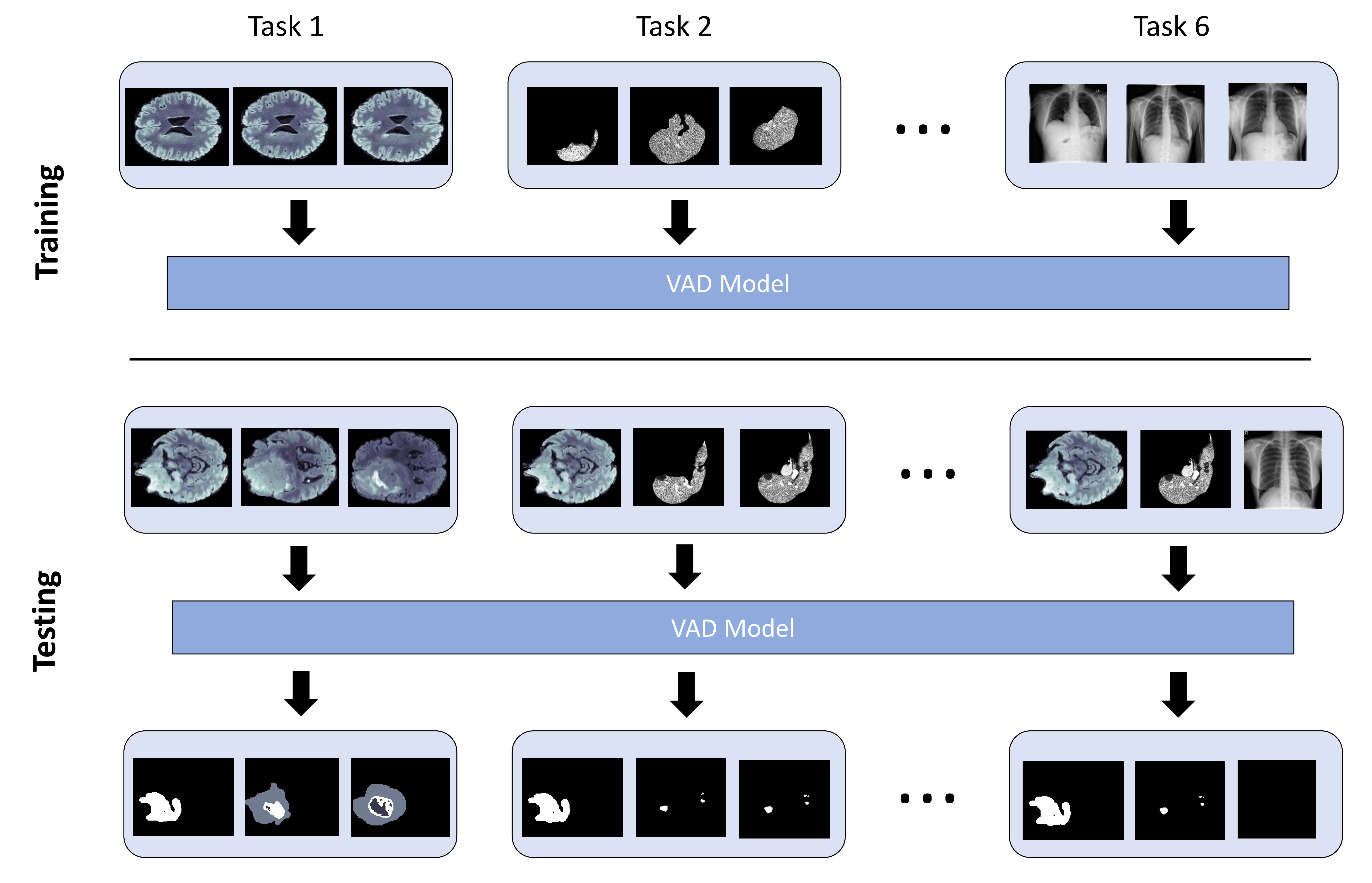}
  \caption{Continual Learning Visual Anomaly Detection for the BMAD medical imaging dataset.}
  \label{fig:teaser}
\end{figure}

\section{Related Work}
\label{sec:related}

\subsection{Visual Anomaly Detection}
\label{subsec:related_work_vad}

A lot of research has been conducted in the VAD domain during the last years, and several VAD models have been developed.
Most VAD methods fall into reconstruction-based and feature embedding-based methods. \\
\textbf{Reconstruction-based methods} use generative models to learn normal image reconstruction during training: large reconstruction errors during inference indicate anomalies. Approaches like AutoEncoders, GANs and Diffusion Models are prominent in this area. An example of these models are \cite{draem} and \cite {riad}. \\
\textbf{Feature-based methods} rely on data representations generated by pre-trained neural networks. 
For instance, PatchCore is a feature-based method, which saves features of normal images in a memory bank during training to compare them with test images features during inference.

These techniques balance anomaly detection performance and interpretability by providing both image-level and pixel-level scores and visual heatmaps that pinpoint the image areas depicted as anomalies.
\\
In the last years, VAD research has moved from implementing better models in terms of accuracy towards  new challenges such as tackling the computational and memory constraints posed by edge devices \cite{paste} \cite{clad_paste} as well as dynamic environments, which can benefit from continual learning techniques \cite{bugarin2024unveiling}.

\subsection{Continual Learning}
\label{subsec:continual_learning}

Traditional Machine Learning models are trained on fixed datasets, but real-world data often differs from the training data distribution. Continual Learning (CL) enables models to adapt to new data without forgetting previous knowledge. Effective CL methods should minimize forgetting, have low memory usage, and be computationally efficient \cite{bugarin2024unveiling}.
\\
CL techniques are generally categorized into three main approaches. \\
\textbf{Rehearsal-based} approaches, such as Experience Replay \cite{expreplay}, store and reuse past data samples during training. Even if very effective, they are not always applicable, especially in situations when it is not possible to store old data due to data privacy issues or memory limitations. \\\textbf{Regularization-based} approaches introduce constraints to the network weights or loss function penalties to retain knowledge of old tasks, often using parameter importance or distillation techniques \cite{ewc} \cite{lwf}. \\
\textbf{Architecture-based} approaches modify the model structure to preserve prior knowledge \cite{pathnet} \cite{packnet}. \\
Experience Replay and, in general, Rehearsal-based approaches are widely considered the most effective strategies to mitigate Catastrophic Forgetting \cite{latent_replay},  particularly for image classification.
However, the best approach may vary by task. For instance, while rehearsal-based methods excel in image classification problem, distillation-based approaches like LwF may be more suitable for the Object Detection \cite{guan2018learn} problem.

\subsection{Continual Visual Anomaly Detection}

In recent years, some interest has been shown in the development of Continual Learning Visual Anomaly Detection (CLAD) models.
One of the first studies proposed a method called Distribution of Normal Embeddings (DNE) \cite{li2022towards}, based on the feature distribution of normal training samples from past tasks.
However, this method operates only at the image level, limiting its relevance in VAD, where pixel-level information, provided by all the current state-of-the-art approaches, is extremely relevant for interpretability and improving the decision-making process of humans.
\\
Afterwards, \cite{pezze2022continual} evaluates classic VAD models able to provide both image-level and pixel-level information.
This work adapts classic VAD methods by using the replay technique (and other more efficient compressed variants) to make them work in the CLAD scenario. 
Subsequently, \cite{bugarin2024unveiling} expands the previous work by testing several state-of-the-art VAD approaches with the replay technique for many models, such as STFPM \cite{stfpm}, EfficientAD \cite{effad}, and FastFlow \cite{fastflow}, showing different forgetting levels based on the chosen architecture.
\\
In particular, for PatchCore, the replay technique is not applicable since their weights are frozen.
Therefore, the authors introduced an adaptation to the model to enable it to evolve over time while retaining previous knowledge, while at the same time ensuring that the memory size remains fixed, preventing unbounded memory growth.
While other recent works were conducted in the field, like \cite{liu2024unsupervised}, all these studies were conducted by considering industrial datasets such as the MVTec dataset \cite{mvtec}.

Though this is a standard practice for the VAD field, focusing only on this specific domain precludes the generalization towards other fields such as the medical field, which could benefit greatly from an adaptive model that can handle new data without the need of retraining.
For this reason, in this work we study for the first time the Continual Visual Anomaly Detection problem in the medical domain by considering the BMAD \cite{bao2023bmad} dataset.

\begin{figure*}[!th]
    \centering
    \includegraphics[width=\textwidth]{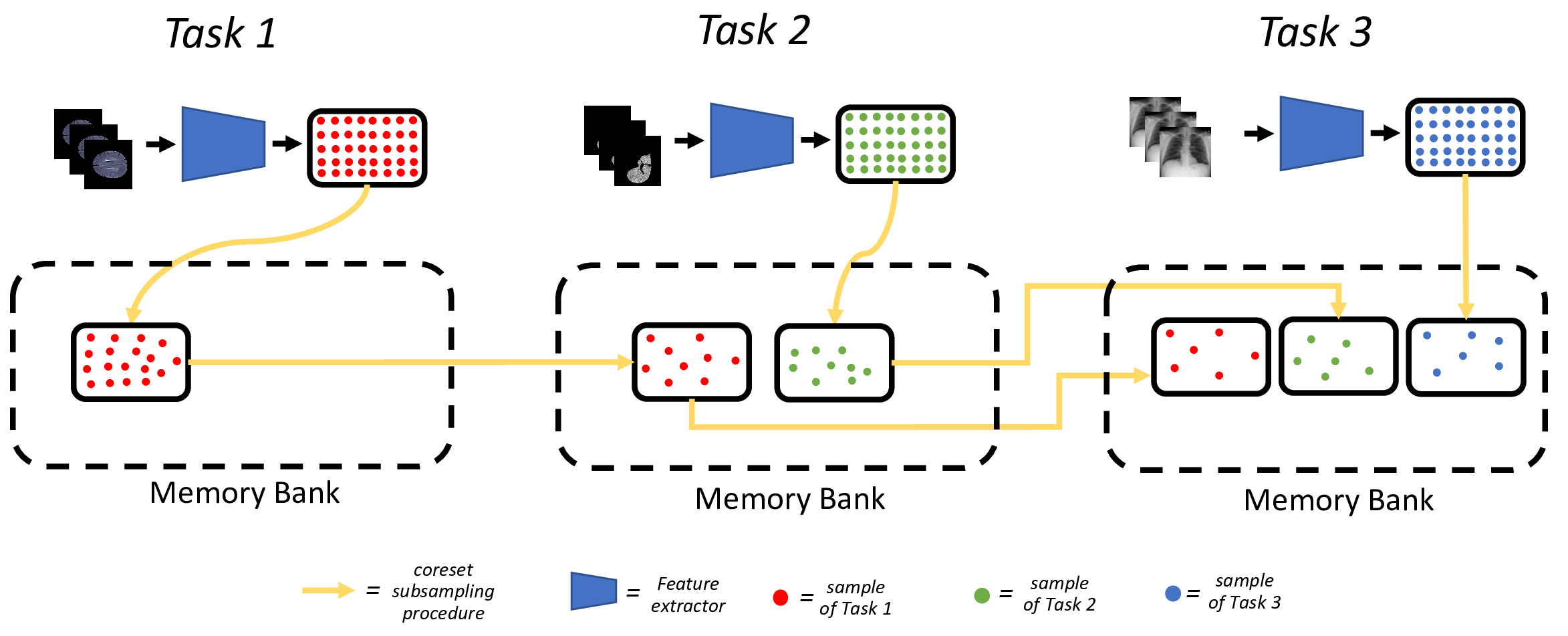}
    \caption{\textbf{Scheme of PatchCoreCL}. When new data arrives, PatchCoreCL applies the procedure of coresubset sampling on the new data and creates a memory bank for that task. At the same time, the coresubset sampling is also applied to the old memory banks to reduce the number of samples. This allows the number of samples stored in memory doesn't change over time.}
    \label{fig:clpatch}
\end{figure*}

\section{Methodology}
\label{sec:method}

\subsection{Preliminaries}
PatchCore \cite{patchcore} is a proven and well known method for Visual Anomaly Detection. This method works by considering a CNN network (such as the WideResnet50 \cite{wideresnet}) as a feature extractor and a memory bank. 
The feature extractor is used for extracting the patch feature vectors from different extraction layers in order to get fine and coarse feature details from the training and test input images. 
The input images of training are thus processed inpatches,s and the memory bank is constructed by storing the most relevant patches.
In particular, during training, the patch feature vectors from the input training images are extracted, and only a subset of them is stored inside the memory bank by considering a KCenter clustering algorithm (called \textbf{coreset subsampling} in the original paper). 
During inference, the patch feature vectors of the input images are compared with a minimax algorithm with the vectors stored in the memory bank. From this patch level anomaly score, it is possible to derive an image-level and a pixel-level anomaly score for the tested images.

\subsection{Continual Medical Anomaly Detection Scenario}
In the Continual Learning scenario, the model is trained in a sequential way by considering a stream of tasks, where each task is associated with a different dataset category (as illustrated in Figure \ref{fig:teaser}). During inference, the model is tested on every category already encountered during the training up to that point. \\
This work considers the BMAD dataset \cite{bao2023bmad}, a collection of six medical datasets from five different domains for medical anomaly detection. BMAD spans multiple anatomical regions, with six different anomaly detection categories: "Brain\_AD", "Liver\_AD", "Retina\_RESC\_AD", "Chest\_AD", "Histopathology\_AD" and \\ "Retina\_OCT2017\_AD".
BMAD is not inherently built as a continual learning dataset, and, in order to test the continual learning strategies, the dataset is split into a sequence of tasks based on the different medical categories, simulating a continual learning scenario. 
Each task includes a binary classification between healthy and anomalous samples, both image-level and pixel-level.\\
Since the task at hand is Visual Anomaly Detection in an unsupervised setting, the training dataset contains only normal samples, while, at test time, both normal (unseen during training) and abnormal samples are evaluated.
The stream of tasks is composed of the following categories: "Brain\_AD", "Liver\_AD", "Retina\_RESC\_AD", "Chest\_AD", "Histopathology\_AD" and "Retina\_OCT2017\_AD".

\subsection{Continual PatchCore (PatchCoreCL)}

In this work, we consider PatchCore as model for our study.
However, given its static nature, classic CL methods like replay cannot work.
Indeed, the PatchCore model keeps its weights (the feature extractor) frozen, and only the memory bank containing patches is updated just once using the training data.
\\
Therefore, it is necessary to develop strategies that enable the PatchCore model to work in incremental scenarios.
In particular, our study implements the idea proposed in \cite{bugarin2024unveiling} which we refer with the name of \textit{PatchCoreCL} and allows PatchCore to work in the continual learning setting.
\\
To adapt the PatchCore model to the dynamic continual learning scenario, several modifications were made to its internal memory bank and training procedure. The core idea is to maintain a separate memory bank for each task encountered during training while keeping the total number of stored patch feature vectors fixed to avoid problems of exploding memory.
This means that over time, the number of associated vectors for a task is given by the maximum number of samples that can be stored in the memory bank divided by the number of tasks encountered. 
In practice, this translates in a ever-decreasing number of samples associated to each task.
However, instead of applying random sampling on the patches, PatchCoreCL applies a memory bank update procedure that applies separetely the coreset reduction step on the stored samples for each task.
The training procedure is illustrated in Algorithm \ref{patchcore_training} while the whole procedure is shown in Figure \ref{fig:clpatch}.

\begin{algorithm}
\caption{PatchCore memory bank update for CL method}
\label{patchcore_training}
\begin{algorithmic}[1]
\State \textbf{Initialize:} $\mathcal{M} \gets$ empty list
\State $memory\_size \gets 30000 \space | \space 10000$ 
\For{$i = 0$ to $N - 1$} \Comment{$N$ = number of tasks}
    \State $p \gets$ patches extracted for task $i$
    \For{$j = 0$ to $length(\mathcal{M})$}
        \State $\mathcal{M}[j] \gets \textit{coreset\_subsampling}(\mathcal{M}[j], \frac{memory\_size}{i+1})$
    \EndFor
    \State $m \gets \textit{coreset\_subsampling}(p, \frac{memory\_size}{i+1})$
    \State $\mathcal{M}.append(m)$
\EndFor
\end{algorithmic}
\end{algorithm}

Within the realm of Continual Learning, this methodology is based on a rehearsal strategy for adapting the model to the new tasks, since a portion of old data (or more precisely, portions of the feature maps associated with the images) is stored in the memory banks.
During inference, when a sample arrives, the patch feature vectors are compared with every memory bank in the list in order to get an image-level score. 
The memory bank that ensures the lowest score is then used for calculating the pixel-level score and the final anomaly mask. 
This is reasonable because the memory banks associated with categories different from the test sample category will cause a higher image-level anomaly score. 
Thanks to this inference strategy, this Continual Learning adaptation is also able to perform task identification during inference: a critical phase for every continual learning technique.

\label{fig:f1_over_tasks}
\begin{figure}[!ht]
    \centering
    \includegraphics[width=0.65\columnwidth]{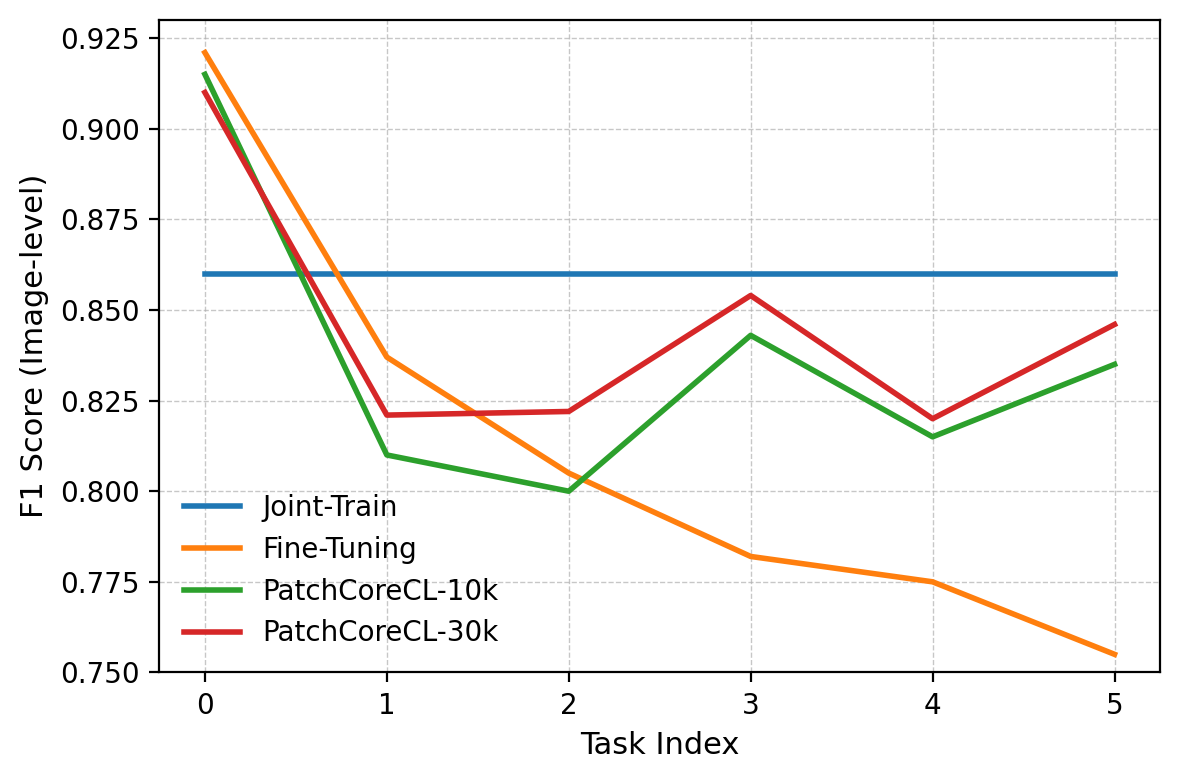}
    \caption{Image-level F1 score for all the tested methods, the task index is chronologically ordered to match the task stream in the continual learning procedure. The F1 is computed for each method as the mean score for all the tasks seen in training up to that point.}
    \label{fig:your_figure_label}
\end{figure}

\section{Experimental Setting}
\label{sec:exp_setting}

\subsection{Dataset}

As described in Section \ref{sec:method}, this work considers the BMAD dataset in a Continual Learning scenario, where the stream of tasks is:  "Brain\_AD", "Liver\_AD", "Retina\_RESC\_AD", "Chest\_AD",
"Histopa- thology\_AD" and "Retina\_OCT2017\_AD". The first three categories are labeled at both image and pixel levels, while the last three are labeled only at the image level. 
\\
The images are considered at a resolution of 224 x 224 pixels \\
Furthermore, due to the large number of samples present for every BMAD category, we decided to consider a maximum of 2000 samples per category for training.

\subsection{Compared Strategies}

We compare PatchCore under various scenarios to assess its performance in continual learning and compare it with relevant baselines. 

\textit{Multi-Model:} Independently trains one model per task. This is the standard approach in the literature and serves as an upper bound reference, but does not support any form of knowledge transfer or continual adaptation.

\textit{Joint-Train:} A single model is trained jointly on all tasks. While not a continual method, it offers a performance upper bound assuming full access to all data at once.

\textit{Fine-Tuning:} A baseline continual learning setup where the model is updated sequentially with no mitigation of forgetting effects. It illustrates the degradation from naive updates.

\textit{PatchCoreCL:} The proposed PatchCore adaptation to the CL scenario, described in Section \ref{sec:method}. The tested versions are:
\begin{itemize}
    \item PatchCoreCL-30K: considers a memory bank of 30000 patch feature vectors.
    \item PatchCoreCL-10K: considers a memory bank of 10000 patch feature vectors.
\end{itemize}

\section{Metrics}
\label{sec:metrics}

We evaluate performance across both anomaly detection and continual learning dimensions using standard metrics. Results are reported for image-level and pixel-level anomaly detection, as well as for continual learning-specific metrics.

\subsection{Anomaly Detection Metrics}

We first assess the anomaly detection capabilities of each strategy using standard Visual Anomaly Detection metrics at both the image and pixel levels.
\\
\textit{Image-level metrics} assess the model’s ability to classify an entire image as anomalous or not. We report the Area Under the Receiver Operating Characteristic Curve (AUROC) and F1 score based on the image-level anomaly score.
\\
\textit{Pixel-level metrics} evaluate the spatial localization of anomalies via the heatmaps produced by PatchCore. Applicable only for subsets of BMAD that include ground truth segmentation masks, we compute:
\begin{itemize}
    \item AUROC and F1 score based on per-pixel anomaly scores,
    \item Precision-Recall (PR) score,
    \item AU-PRO, the Area Under the Per-Region Overlap curve, which measures overlap between predicted and ground truth anomalous regions.
\end{itemize}

In the results obtained in the CL scenario, we report all the image-level metrics as the average across all tasks, measured after the model has completed training on the final task.
BMAD spans multiple anatomical regions, with six different anomaly detection categories, three of which support pixel-level evaluation ("Brain\_AD", "Liver\_AD" and "Retina\_RESC\_AD") while the remaining three are for sample-level assessment only.
Therefore, the pixel-level metrics are obtained after training the model on the first three tasks, since only the first three categories in the stream are pixel-level labeled. Furthermore, for the same reason, the relative gap and the average forgetting are calculated at the pixel level.

\subsection{Continual Learning Metrics}

To assess continual learning performance, we compute:

\textit{Relative Gap ($\delta$)}: For each continual learning method, we measure the difference between its performance pixel-level F1 ($\text{F1}_{\text{CL}}$) and the Joint-Train upper bound ($\text{F1}_{\text{Joint}}$). 

Defined as $\delta$:
$$
%\delta = \max(\text{F1}_{\text{Multi}}, \text{F1}_{\text{Joint}}) - \text{F1}_{\text{CL}}
\delta = \text{F1}_{\text{Joint}} - \text{F1}_{\text{CL}}
$$

\textit{Average Forgetting}: Mean performance drop across all previously seen tasks when training on a new task. This quantifies the extent to which past knowledge is overwritten in the continual setting. This metric quantifies how much of the previous task knowledge is lost during training.

Defined as $F$:
$$
F = \frac{1}{T-1} \sum_{t=1}^{T-1} \left( \max_{k \in \{1, \dots, t\}} R_{k,t} - R_{T,t} \right)
$$
\indent where:
\begin{quote}
\begin{itemize}
    \item[$R_{k,t}$] is the performance (e.g., F1 score) on task $t$ after training on task $k$,
    \item[$R_{T,t}$] is the performance on task $t$ after training on the final task $T$.
\end{itemize}
\end{quote}

In this work, we report the Average Forgetting values as percentage values, as done in \cite{bugarin2024unveiling}.

When evaluating Continual Learning strategies, it is very important to evaluate the relative gap and the average forgetting at the same time.  This comparison is essential, as a model may exhibit low forgetting but substantially lower performance compared to the Joint-Train strategy, which is the upper bound for all the CL approaches.

In addition, for every CL strategy, we reported the Architecture Memory (needed for storing the PatchCore feature extractor) and the Additional Memory (needed for storing the Memory Bank) in terms of MegaBytes (MB).
In Continual Learning we are interested in updating and expanding the model's knowledge over time, avoiding forgetting, while at the same time, we want to obtain this goal with minimal computation and memory overhead. 
Therefore, these additional metrics are extremely relevant for assessing the practical feasibility and deployment of Continual Learning models in real-world environments.

\begin{table}[!ht]
\centering
\caption{
    Comparison of PatchCore performance across different learning strategies on the BMAD benchmark. Metrics are reported at both image-level and pixel-level, including AUROC, F1, precision-recall (PR), and AUPRO. We compare Multi-Model, Joint-Train, and the continual learning strategies (Fine-Tuning, PatchCoreCL-10k, PatchCoreCL-30k). The relative gap to the upper bound and average forgetting are also reported to assess performance in continual learning. We also report the Architecture and Additional Memory.
}
\label{tab:results}
\begin{tabular}{ll|c|c|c|c|c}
\toprule
\multicolumn{2}{c|}{\multirow{2}{*}{\textbf{PatchCore}}} &
  \multirow{2}{*}{\makecell{\textbf{Multi} \\ \textbf{Model}}} &
  \multirow{2}{*}{\makecell{\textbf{Joint} \\ \textbf{Train}}} &
  \multirow{2}{*}{\makecell{\textbf{Fine} \\ \textbf{Tuning}}} &
  \multicolumn{2}{c}{\textbf{PatchCoreCL}} \\
\multicolumn{2}{c|}{} &
  \multicolumn{1}{c|}{} &
  \multicolumn{1}{c|}{} &
  \multicolumn{1}{c|}{} &
  \multicolumn{1}{c|}{\textbf{10k}} &
  \textbf{30k} \\
\midrule
\multicolumn{1}{l}{
    \multirow{2}{*}{\makecell[l]{\textbf{Image} \\ \textbf{level}}}
} &
  \multicolumn{1}{l|}{AUROC} &
  0.99 &
  0.82 &
  0.73 &
  \multicolumn{1}{c|}{0.77} &
  0.81 \\ 
\multicolumn{1}{l}{} &
  \multicolumn{1}{l|}{F1} &
  0.98 &
  0.86 &
  0.75 &
  \multicolumn{1}{c|}{0.84} &
  0.85 \\
\midrule
\multicolumn{1}{l}{} &
  \multicolumn{1}{l|}{AUROC} &
  0.94 &
  0.93 &
  0.31 &
  \multicolumn{1}{c|}{0.93} &
  0.93 \\ 
\multicolumn{1}{l}{\multirow{2}{*}{\makecell[l]{\textbf{Pixel} \\ \textbf{level}}}} &
  \multicolumn{1}{l|}{F1} &
  0.64 &
  0.63 &
  0.25 &
  \multicolumn{1}{c|}{0.62} &
  0.63 \\ 
\multicolumn{1}{l}{} &
  \multicolumn{1}{l|}{PR} & % PR  = Precision Recall
  0.70 &
  0.69 &
  0.12 &
  \multicolumn{1}{c|}{0.67} &
  0.69 \\ 
\multicolumn{1}{l}{} &
  \multicolumn{1}{l|}{AUPRO} &
  0.78 &
  0.76 &
  0.10 &
  \multicolumn{1}{c|}{0.75} &
  0.77 \\
\midrule
\multicolumn{2}{l|}{\textbf{Arch. Mem \hfill [MB]}} &
  275.6 &
  275.6 &
  275.6 &
  \multicolumn{1}{c|}{275.6} &
  275.6 \\ 
\multicolumn{2}{l|}{\textbf{Add. Mem \hfill [MB]}} &
  1105.8 &
  184.3 &
  184.3 &
  \multicolumn{1}{c|}{61.43} &
  184.3 \\ 
\multicolumn{2}{l|}{\textbf{Rel. Gap ($\delta$)}} &
  - &
  - &
  0.11 &
  \multicolumn{1}{c|}{0.02} &
  0.01 \\ 
\multicolumn{2}{l|}{\textbf{Avg. Forget. \hfill [\%]}} &
  - &
  - &
  52.32 &
  \multicolumn{1}{c|}{0.80} &
  0.73 \\
\bottomrule
\end{tabular}%
\end{table}

\section{Results}
\label{sec:results}

This section analyzes the results obtained with the different strategies listed in Section \ref{sec:exp_setting}.
In Table \ref{tab:results} all the obtained Anomaly Detection and Continual Learning metrics are reported, along with performance metrics such as additional memory. 

As reported in Table~\ref{tab:results}, the best performing strategy is Multi-Model as expected, where a different model is trained for each BMAD category. This solution is unsurprisingly the best in terms of accuracy, both image- and pixel-level, but implies a large additional memory consumption (1105.8 MB), needed for storing all extracted memory banks (one for each task) while only a single feature extractor is stored. 
\\
The Joint-Train strategy, which defines the upper bound for a single-model performance, allows good results at both image- and pixel-level with a reduced memory usage. 
Instead, the Fine-Tuning approach, as always reported in the literature, is the worst strategy. It achieves the lower results in terms of both image- and pixel-level accuracies. 
\\
The two proposed sample-based strategies achieved very promising results, reaching almost the Joint-Train performances on both image- and pixel-level metrics. On image-level metrics, PatchCoreCL-10k suffers a GAP of 0.02 on the F1 score and only 0.01 on the F1 pixel-level, while the PatchCoreCL-30k replicates perfectly the Joint-Train performances. 
\\
Very interesting is PatchCoreCL-10k, which uses only one-third of the memory required by Joint-Train, and the obtained performances are not too far from it, with a relative gap of only 0.02 and a very contained forgetting (0.80\% against the 0.73\% of the PatchCoreCL-30K). The last results show how the chosen memory bank update strategy allows for reducing task interference between tasks and allows for retaining the past  knowledge in a very effective way.

In Figure \ref{fig:f1_over_tasks}, the trend of the F1 image-level metric among the considered tasks is reported. As can be visually seen, the Fine-Tuning solution shows the highest forgetting, while the PatchCoreCL-10k and PatchCoreCL-30k solutions show a very small forgetting, a lower gap to the Joint-Train approach and a good retention of the previous knowledge.

\section{Conclusions}
\label{sec:conclusions}

In this work, we explored for the first time the Visual Anomaly Detection problem in a Continual Learning scenario for the medical domain, where new domains are incrementally presented to the VAD model.
\\
Our results are tested on a CL scenario created from the well-established BMAD dataset.
As the main CL technique, we considered the PatchCoreCL method, providing a robust baseline for future research.
\\
The results show how PatchCoreCL-30K and PatchCoreCL-10K can tackle the problem in a very effective way, showing a very small relative gap with respect to the Joint-Train solution and very low forgetting.
Specifically, PatchCoreCL-30K obtains a forgetting value less than 1\%.
This is accomplished by considering the same amount of memory used in the Joint-Train or even less memory (one-third) in the PatchCoreCL-10K solution.
This work paves the way for the development of new Visual Anomaly Detection models under this particular and challenging scenario. Possible future work directions can be focused on improving even more the proposed PatchCore adaptation and on testing even more state-of-the-art Visual Anomaly Detection models on this configuration.

\section*{Acknowledgments}
We thank Prof. Damiano Varagnolo for his valuable insights and constructive feedback.
\\
Moreover, this study was also partially carried out within the PNRR research activities of the consortium iNEST (Interconnected North-Est Innovation Ecosystem) funded by the European Union Next-GenerationEU (Piano Nazionale di Ripresa e Resilienza (PNRR) - Missione 4 Componente 2, Investimento 1.5 - D.D. 1058 23/06/2022, ECS00000043).

\bibliographystyle{unsrt}  
\bibliography{references}

\end{document}